\documentclass{article}

\usepackage[preprint]{neurips_2023}

\usepackage{natbib}
\setcitestyle{numbers,square}

\usepackage[utf8]{inputenc} % allow utf-8 input
\usepackage[T1]{fontenc}    % use 8-bit T1 fonts
\usepackage{hyperref}       % hyperlinks
\usepackage{url}            % simple URL typesetting
\usepackage{booktabs}       % professional-quality tables
\usepackage{amsfonts}       % blackboard math symbols
\usepackage{nicefrac}       % compact symbols for 1/2, etc.
\usepackage{microtype}      % microtypography
\usepackage{xcolor}         % colors

\usepackage{times}
\usepackage{epsfig}
\usepackage{graphicx}
\usepackage{amsmath}
\usepackage{amssymb}

\usepackage{amsmath}
\usepackage{amssymb}
\usepackage{mathtools}
\usepackage{amsthm}

\usepackage{multirow}
\usepackage{algorithm}
\usepackage{algorithmic}

\usepackage{xcolor}
\usepackage{xspace} % for use in macros
\definecolor{purple}{rgb}{0.2, 0, 1.0}
\definecolor{darkred}{rgb}{0.8, 0.1, 0.1}
{}  % {new text}{old text}
{}

\title{AttentionX: Exploiting Consensus Discrepancy In Attention from A Distributed Optimization Perspective} 

\author{%
  Guoqiang Zhang  \\
  Department of Computer Science \\
  University of Exeter, UK \\
  \texttt{g.z.zhang@exeter.ac.uk} 
  \\
   \And  Richard Heusdens \\ 
  Netherlands Defence Academy (NLDA) \\
  The Netherlands \\
 \texttt{r.heusdens@\{mindef.nl, tudelft.nl\}} \\
}

\begin{document}

\maketitle
\begin{abstract}
%One popular diffusion-based sampling strategy attempts to solve the reverse ordinary differential equations (ODEs) effectively. 
In this paper, we extend the standard Attention in transformer by exploiting the consensus discrepancy from a distributed optimization perspective, referred to as AttentionX. It is noted that %the  popular distributed optimization algorithm %ADMM \cite{Boyd11ADMM} and
the primal-dual method of multipliers (PDMM) \cite{Zhang16PDMM} is designed to iteratively solve a broad class of distributed optimization problems over a pear-to-pear (P2P) network, where neighbouring nodes gradually reach consensus as specified by predefined linear edge-constraints in the optimization process. In particular,  at each iteration of PDMM, each node in a network first performs information-gathering from neighbours and then performs local information-fusion. From a high-level point of view, the $KQ$-softmax-based weighted summation of $V$-representations in Attention corresponds information-gathering from neighbours while the feature-processing via the feed-forward network (FFN) in transformer corresponds to local information fusion. PDMM exploits the Lagrangian multipliers to capture the historical consensus discrepancy in the form of residual errors of the linear edge-constraints, which plays a crucial role for the algorithm to converge. Inspired by PDMM, we propose AttentionX to incorporate the consensus discrepancy in the output update-expression of the standard Attention. The consensus discrepancy in AttentionX refers to the difference between the weighted summation of $V$-representations and scaled $V$-representions themselves. Experiments on ViT and nanoGPT show promising performance.

\end{abstract}

%%%%%%%%%%%%%%%%%%%%%%%%%%%%%%%%
\section{Introduction}
\label{sec:intro}
%%%%%%%%%%%%%%%%%%%%%%%%%%%%%%%%

In recent years, transformers \cite{Transformer17} in artificial intelligence have achieved great success in various data-analysis domain such as  natural language processing (NLP) \cite{Achiam23GPT34,Touvron23Llama2}, computer vision \cite{Dosovitskiy21ViT}, image generation and editing \cite{Peebles23ViTDiffusion, Hatamizadeh24DiffiT,Aida23BDIA}, and audio processing \cite{Latif23audioTrans}. One key component in transformer is the Attention layer for capturing long-distance dependency across a sequence of tokens. Specifically, the Attention operation gathers relevant information from other tokens for each one via the $KQ$-softmax-based weighted summation of $V$-representations. The feedforward network (FFN) further processes the output of the Attention layer per token, which can be interpreted as local information fusion.  

One bottleneck of the standard Attention is that its computational complexity is quadratic in terms of the number of tokens, which becomes infeasible for extremely long sequences of tokens. As a result, various simplified Attention schemes have been proposed to reduce the complexity of the  standard Attention, which include, for example, LinFormer \cite{Wang20LinFormer}, LongFormer \cite{Beltagy20LongFormer}, ReFormer \cite{Kitaev20Reformer},  FlashAttention \cite{Dao23Falshattion}, RingAttention \cite{Liu23RingAtten}, BurstAttention\cite{Sun24BurstAttention}. We note that all the above schemes intend to reduce computational complexity instead of improving performance from a distributed optimization perspective.       

From a high-level point of view, the Attention-FFN framework in transformers exhibits a certain similarity with the framework of distributed optimization of which typical algorithms include alternating direction method of multipliers (ADMM) \cite{Boyd11ADMM} and primal-dual method of multipliers (PDMM) \cite{Zhang16PDMM}. Considering PDMM as an example, it was primarily designed to solve the following separable convex optimisation problem 
\begin{align}
\begin{array}{ll} \text{minimise} & {\displaystyle \sum_{i\in {\cal V}} f_i(x_i)} \\\rule[4mm]{0mm}{0mm}
\text{subject to} & A_{ij}x_i + A_{ji}x_j = b_{ij}, \quad (i,j)\in \cal  E, \label{equ:linearCond}
\end{array}
\end{align} 
where the undirected graph $G=(\mathcal{V},\mathcal{E})$ represents a pear-to-pear (P2P) network from practice, and  each node $i$ carries a local objective function $f_i(\cdot)$ and each edge $(i,j)$ carries a linear equality constraint as specified by the constant $(A_{ij}, A_{ji}, b_{ij})$.  As will be discussed in detail in Section~\ref{sec:PDMM}, at each iteration of PDMM, each node in the network performs local information gathering from neighbours (corresponding to Attention in transformer) and local information fusion (corresponding to FFN). One key property of PDMM is that its update expression utilizes the consensus discrepancy in terms of the residual error of the linear edge-constraints in (\ref{equ:linearCond}), which is essential to make the algorithm converge.  

In this paper, we aim to extend the standard Attention by following a similar procedure as PDMM. In particular, we compute the consensus discrepancy in Attention as the difference between the scaled weighted summation of the $V$-representations and the $V$-representations themselves. It is hypothesized that as the index of the Attention-FFN layer increases, the computed consensus discrepancy tends to decrease and become stable as the dependency across all the tokens would be fully captured when the transformer gets very deep. This is conceptually similar to the consensus achieved by PDMM to a certain extent, of which the residual errors of the linear edge-constraints gradually decreases as the iteration of PDMM increases. Similarly to PDMM, we propose to incorporate the above consensus discrepancy in computing the output of Attention, referred to as AttentionX. Experiments indicate that AttentionX indeed improves the validation performance of ViT and nanoGPT.   

We note that a recent work \cite{Ye24DiffTransformer} (which was made public after our work) proposes to calculate the attention scores as the difference between two separate softmax attention maps before multiplying the $V$-representations. The purpose for doing so is to amplify attention to the relevant context
while canceling noise. Even though our work and \cite{Ye24DiffTransformer} have different motivations, from a high-level point of view, both approaches attempt to calculate the difference of two separate weighted summations of $V$-representations. 

\section{Brief Review of PDMM}
To facilitate node-oriented distributed optimization of (\ref{equ:linearCond}) over a graph $G=(\mathcal{V},\mathcal{E})$, PDMM introduces two Lagrangian multipliers $\lambda_{i|j}$ and $\lambda_{j|i}$ for each linear constraint over the edge $(i,j)\in \mathcal{E}$.  Let $\mathcal{N}_i$ denote the set of neighbors for node $i$. At the $k$th iteration, each new update $\boldsymbol{x}_i^{k+1}$ is computed in terms of the information $\{(x_{j|i}^{k}, \lambda_{j|i}^k)| j\in \mathcal{N}_i\}$   from neighbors as  
\label{sec:PDMM}
\begin{align}
x_i^{k+1} &= \arg\min_{x_i} 
\underbrace{\left[ f_i(x_i) - x_i^T\overbrace{(\sum_{i\in \mathcal{N}_i}A_{ij}^T\lambda_{j|i}^k)}^{\textrm{\textcolor{blue}{1st info. gathering }} %\\\textrm{\textcolor{blue}{(1st term)}} 
} + \overbrace{\sum_{j\in \mathcal{N}_i} \frac{\rho}{2}\|A_{ij} x_i+A_{ji}x_j^{k}-b_{ij}\|^2}^{\textrm{\textcolor{blue}{2nd info. gathering}}}
\right]}_{\textrm{\textcolor{blue}{info. fusion}}} \quad \forall i\in \mathcal{V}, \label{equ:PDMM_x} 
\end{align}
where the scalar $\rho>0$. Once $x_i^{k+1}$ is available, the associated Lagrangian multipliers of node $i$ are updated to be 
\begin{align}
\lambda_{i|j}^{k+1} &= \lambda_{j|i}^k + \rho (b_{ij}-A_{ji}x_j^k-A_{ij}x_i^{k+1}) \quad \forall i\in \mathcal{V}\textrm{, } j\in \mathcal{N}_i.
\label{equ:PDMM_lambda}
\end{align}
Detailed convergence results of the algorithm can be found in \cite{Zhang16PDMM, Sherson17PDMM}.   

By inspection of (\ref{equ:PDMM_x}), it is seen that the computation of $x_i^{k+1}$ involves two weighted summations from neighbors, which are $\sum_{i\in \mathcal{N}_i}A_{ij}^T\lambda_{j|i}^k$ and $\sum_{j\in \mathcal{N}_i}A_{ij}^T(A_{ji}x_j^k-b_{ij})$ as contributed by the first and second information gathering terms. $x_i^{k+1}$ is then obtained by solving a small-size optimization problem with the local function $f_i(\cdot)$, and can be viewed as local information fusion.  

Next we study the Lagrangian multiplier $\lambda_{j|i}^k$ being explored in the computation of $x_i^{k+1}$. It is not difficult to conclude from (\ref{equ:PDMM_lambda}) that $\lambda_{j|i}^k$ can be represented as a summation of the historical residual errors of the linear equality constraint for edge $(i,j)\in \mathcal{E}$. For the case of $k$ being even, $\lambda_{j|i}^k$ can be represented as  
\begin{align}
%\lambda_{j|i}^k &= \lambda_{i|j}^{k-1} + \rho (b_{ij}-A_{ji}x_j^{k-1}-A_{ij}x_i^{k}).  \\
%& =\lambda_{j|i}^{k-2} +\rho (b_{ij}-A_{ji}x_j^{k-2}-A_{ij}x_i^{k-1}) + \rho (b_{ij}-A_{ji}x_j^{k-1}-A_{ij}x_i^{k}).  \\
& =\lambda_{j|i}^{0} +\rho \sum_{m=1}^{k/2}  (b_{ij}-A_{ji}x_j^{2m-2}-A_{ij}x_i^{2m-1}) + \rho \sum_{m=1}^{k/2} (b_{ij}-A_{ji}x_j^{2m-1}-A_{ij}x_i^{2m}).  
\label{equ:lambda_his}
\end{align}
We take each residual error in (\ref{equ:lambda_his}) as the measurement of the consensus discrepancy between the pair of nodes $(i,j)$.   

In addition to the Lagrangian multipliers for capturing the historical consensus discrepancy, it is clear from (\ref{equ:PDMM_x}) that the set of quadratic penalty functions $\{\|A_{ij}x_i+A_{ji}x_j^k-b_{ij}\|\}_{j\in \mathcal{N}_i}$ are also included in computation of $x_i^{k+1}$. The penalty functions attempt to softly constrain $x_i^{k+1}$ in a region that incurs small consensus discrepancy (with regard to the predefined edge-constraints) with respect to the neighbors $\{x_j^k\}_{j\in \mathcal{N}_i}$. The parameter $\rho>0$ in front of the penalty functions and in (\ref{equ:lambda_his}) controls the  contribution of the consensus discrepancy when updating the primal variables $\{x_i\}_{i\in \mathcal{V}}$.  

Since the invention of PDMM, it has received considered research investigation in the past few years. The work \cite{jor:23} studied the convergence of stochastic PDMM which includes  asynchronous PDMM and PDMM with transmission losses between neighbours as special cases.  In \cite{zha:22}, PDMM is modified for federated learning over a centralised network, where it is found that PDMM is closely related to the SCAFFOLD \cite{kar:20} and FedSplit \cite{pat:20} algorithm. Additionally, PDMM can be employed for privacy-preserving distributed optimisation, providing a level of privacy assurance, by utilising the fact that the (synchronous) PDMM updates take place within a particular subspace and the orthogonal complement can be used  to obscure local (private) data, a technique known as subspace perturbation \cite{li:20, li:21, Li:20sp, li:24}. Additionally, research in \cite{jon:18} demonstrates that PDMM exhibits robustness against data quantisation.
Recently, the PDMM algorithm has been extended to incorporate affine inequality constraints as well \cite{heu:24}.
This enhancement enables its application in solving linear programs in a distributed fashion.

\section{AttentionX by Incorporating Consensus Discrepancy} 

\subsection{Revisiting Attention-FFN framework in transformer}

The original work \cite{Transformer17} proposes the encoder-decoder structure in the transformer for NLP applications.  The Attention-FFN framework is slightly different in encoder and decoder. For demonstration purpose, we consider a simplified version, represented as (see \cite{MHA23Pytorch, Dosovitskiy21ViT}) 
\begin{align}
\textrm{head}_m(X) &= \textrm{Attention}(\overbrace{XW_m^{Q}}^{Q_m}, \overbrace{XW_m^{K}}^{K_m}, \overbrace{XW_m^V}^{V_m}) \label{equ:att_ffn1} \\
\textrm{MultiHead}(X) &= \textrm{Concat}(\textrm{head}_1(X),\ldots, \textrm{head}_h(X))W^{o} \label{equ:att_ffn2} \\
Y &= X+\textrm{MultiHead}(X) \label{equ:att_ffn2_5}  \\
X &= \underbrace{\textrm{FFN}(Y)+Y}_{\textrm{\textcolor{blue}{info. fusion}}},
\label{equ:att_ffn3}
\end{align}
where $(W_m^Q, W_m^K, W_m^V)$ are the three learnable matrices for computing $(Q_m, K_m, V_m)$ of the $m$th attention, and Concat stacks up $h$ attentions $\{\textrm{head}_m(X)\}_{m=1}^h$, which is multiplied by the learnable matrix $W^o$ as the output of the Attention layer. 

It is well-known that the attention operation in Equ.~(\ref{equ:att_ffn1}) is a QK-softmax-based weighted summation of $V$ representations, given by 
\begin{align}
\textrm{head}_m(X) = \overbrace{\textrm{softmax}\left(\frac{Q_mK_m^T}{\sqrt{d_m}}\right)V_m}^{\textrm{\textcolor{blue}{info. gathering}}},
\end{align}
where $d_m$ is the dimension of the $Q_m$ vectors. The softmax term computes the unified relevance of each token with respect to neighbouring tokens. Similarly to that of PDMM, the computed weighted summation of $V$ representations can be taken as information gathering from all neighbours.  %It is noted that for windowed attention across time, the  

The update expression (\ref{equ:att_ffn3}) processes the output of multi-head attentions  via FFN and skip-connection on a per-token basis. Therefore, it can be viewed as local information fusion. One difference between (\ref{equ:att_ffn3}) and (\ref{equ:PDMM_x}) of PDMM is that the parameters of FFN are shared by all the tokens while the individual functions $\{f_i(\cdot)|i\in \mathcal{V}\}$ are in general pre-defined and node-dependent.   

\subsection{Update expression of AttentionX} 

As reviewed earlier, PDMM exploited consensus discrepancy in the form of residual errors of the linear equality constraints in its update expressions. Similarly, we extend Attention by also incorporating the associated consensus discrepancy. We let the consensus discrepancy in the context of Attention as the difference between the scaled weighted summation of $V$ representations and $V$ representations themselves, given by 
\begin{align}
\Phi_m(X)= V_m -\gamma\textrm{softmax}\left(\frac{Q_mK_m^T}{\sqrt{d_m}}\right)V_m \quad m=1,\ldots, h, \label{equ:Phi}
\end{align}
where $\gamma\geq 1$. Because of the softmax-based weighted summation, more relevant tokens contribute more to the measured consensus discrepancy.
%We hypothesize that as the transformer goes deeper, the dependency across the tokens would be gradually captured by the softmax operation. As a result, the magnitude of consensus discrepancy $\Phi_i$ would decrease and become stable eventually. 
Inspired by PDMM, we compute the output of the multi-head attention in terms of $\{\Phi_m\}_{m=1}^h$ as 
\begin{align}
\textrm{MultiHead}(X) =& \textrm{Concat}(\Phi_1(X),\ldots, \Phi_h(X))W^o.
\label{equ:attX_ffn} 
\end{align}

We refer to (\ref{equ:att_ffn1}) and (\ref{equ:att_ffn2_5})-(\ref{equ:att_ffn3}), together with the new update expressions (\ref{equ:Phi})-(\ref{equ:attX_ffn}) as AttentionX to differentiate it from the classical Attention. It is clear that no additional learnable parameters are introduced in AttentionX. The computational overhead in (\ref{equ:Phi}) only involves one subtration operation, which is negligible.   

Finally, we briefly explain why the consensus discrepancy from earlier AttentionX layers are not utilized for the computing the output of the current AttentionX layer. We first note that in PDMM, the Lagrangian multipliers $\{\lambda_{j|i}\}_{j\in \mathcal{N}_i}$ accumulate all the associated historical residual errors, which are then used for updating the primal variable $\boldsymbol{x}_i$. Intuitively speaking, this is because the linear edge-constraints are fixed over iterations. The historical residual errors play an role in the current iteration. On the other hand, the dependency across tokens in transformers is dynamically learned by stacking a set of Attention-FFN layers.  There is no explicit and fixed constraint between tokens.  Therefore, in AttentionX, we only employ the most recent consensus discrepancy as specified in (\ref{equ:Phi})-(\ref{equ:attX_ffn}).  

\noindent \textbf{Regarding selection of $\gamma$ parameter in (\ref{equ:Phi})}: We argue that the selection of  the $\gamma$ parameter should depend on if the diagonal elements of the weighting matrix $\textrm{softmax}(Q_mK_m^T/\sqrt{d_m})$ are set to be zero or not.  To simplify notations below, we use $\alpha_{k,j}$ to denote the $(k,j)$th element of the weighting matrix.  %Let us first consider the case of zero diagonal elements. 
Consider the $k$th row of $\Phi_m(X)$ for the $k$th token, which can be re-parameterized as
\begin{align}
\Phi_m(X)[k,:] &= V_m[k,:] -\gamma \sum_{j=1}^n  \alpha_{k,j}V_m[j,:]   \nonumber \\
& = \sum_{j=1}^n \alpha_{k,j} ( V_m[k,:]-\gamma V_m[j,:] ) \nonumber\\
& =  \underbrace{\alpha_{k,k} (1-\gamma) V_m[k,:]}_{\textcolor{blue}{\textrm{1st term}}}+ \underbrace{\sum_{j\neq k} \alpha_{k,j} ( V_m[k,:]-\gamma V_m[j,:] )}_{\textcolor{blue}{\textrm{ 2nd term}}} \label{equ:phi_decom}
\end{align}
where we assume there are $n$ tokens in total. In practice, one has the freedom to set $\alpha_{k,k}$ to be zero or not via the masking technique. By  using  (\ref{equ:phi_decom}) and the fact that $\sum_{j=1}^n \alpha_{k,j}=1$, one can easily show that when $\gamma=1$, there is
\begin{align}
\|\underbrace{\Phi_m(X)[k,:]}_{\gamma=1,\alpha_{k,k}=0} \|^2 > \|\underbrace{\Phi_m(X)[k,:]}_{\gamma=1,\alpha_{k,k}\neq 0} \|^2. \label{equ:phi_mag_diff}
\end{align}
That is, the magnitude of the consensus discrepancy $\Phi_m(X)[k,:]$ decreases when $\alpha_{k,k}$ becomes non-zero without any masking.  

To mitigate the magnitude difference in (\ref{equ:phi_mag_diff}), we recommend to set $\gamma>1$ (or $\gamma=1$) when $\alpha_{k,k}\neq 0$ (or $\alpha_{k,k}= 0$).  
The common practice is that in LLMs (e.g., nano-GPT), the diagonal elements are set to be nonzero by including the contributions from their own representations via the casual-masking.

\section{Experiments}
\label{sec:exp}
We evaluated AttentionX for both ViT-small by utilizing the open-source repository, \footnote{\url{https://github.com/kentaroy47/vision-transformers-cifar10}}  and nano-GPT2 by using the repository. \footnote{\url{https://github.com/karpathy/nanoGPT}} It is found that the AttentionX produces promising performance in both tasks.  

\subsection{On training ViT-small}
In this experiment, we consider training ViT-small over CIFAR10 and CIFAR100. The $\gamma$ parameter in AttentionX (see (\ref{equ:Phi})) was set to $\gamma=1$ as the diagonal elements of $\textrm{softmax}(Q_mK_m^T/\sqrt{d_m})$ are set to zero on purpose. The SET-Adam optimizer was utilized \cite{Guoqiang23SETAdam} in the training process with the configuration $(\eta_0, \beta_1,\beta_2,\epsilon)=(1e-4, 0.9, 0.999, 1e^{-18})$, where $\eta_0$ denotes the initial learning rate. The remaining training setups follow directly from the original open source. Three experimental repetitions were performed per training setup to mitigate the effect of randomness.  

Table~\ref{tab:AttentionX_ViT_small} summarizes the obtained validation accuracy. It is clear that ViT-small with AttentionX  produces considerably better performance than with Attention. This indicates that the consensus discrepancy characterised by $\Phi_m(X)$ in (\ref{equ:Phi}) is a better choice than the softmax-based $V$-representations. %Fig.~\ref{fig:VIT_small} visualize the training and validation curves. It is seen that even though the training loss when using the BDIA training technique is higher, the validation loss improves remarkably. This indicates that the random $\gamma$ variable in BDIA indeed regularizes the ViT-small network properly. 

\begin{table}[h!]
\caption{ \footnotesize{Validation accuracy for training ViT-small over CIFAR10 and CIFAR100}} 
\vspace*{-0.0cm}
\label{tab:AttentionX_ViT_small}
\centering
\begin{tabular}{c|c|c|}
 \cline{2-3} 
 & {\scriptsize ViT-small with Attention}  & {\scriptsize ViT-small with AttentionX } 
 \\ \hline
\multicolumn{1}{|c|}{\scriptsize CIFAR10}  &   {\scriptsize 88.15$\pm$0.55 }  & {\scriptsize \textbf{89.41}$\pm$0.18} 
\\ \hline
\multicolumn{1}{|c|}{\scriptsize CIFAR100}  &   {\scriptsize 61.86$\pm$0.47 }  & {\scriptsize \textbf{64.00}$\pm$0.37} 
\\ \hline
\end{tabular}
\vspace*{-0.3cm}
\end{table}

\begin{figure}[t!]
\centering
\includegraphics[width=110mm]
%{attentionx_compare_pro.eps}
{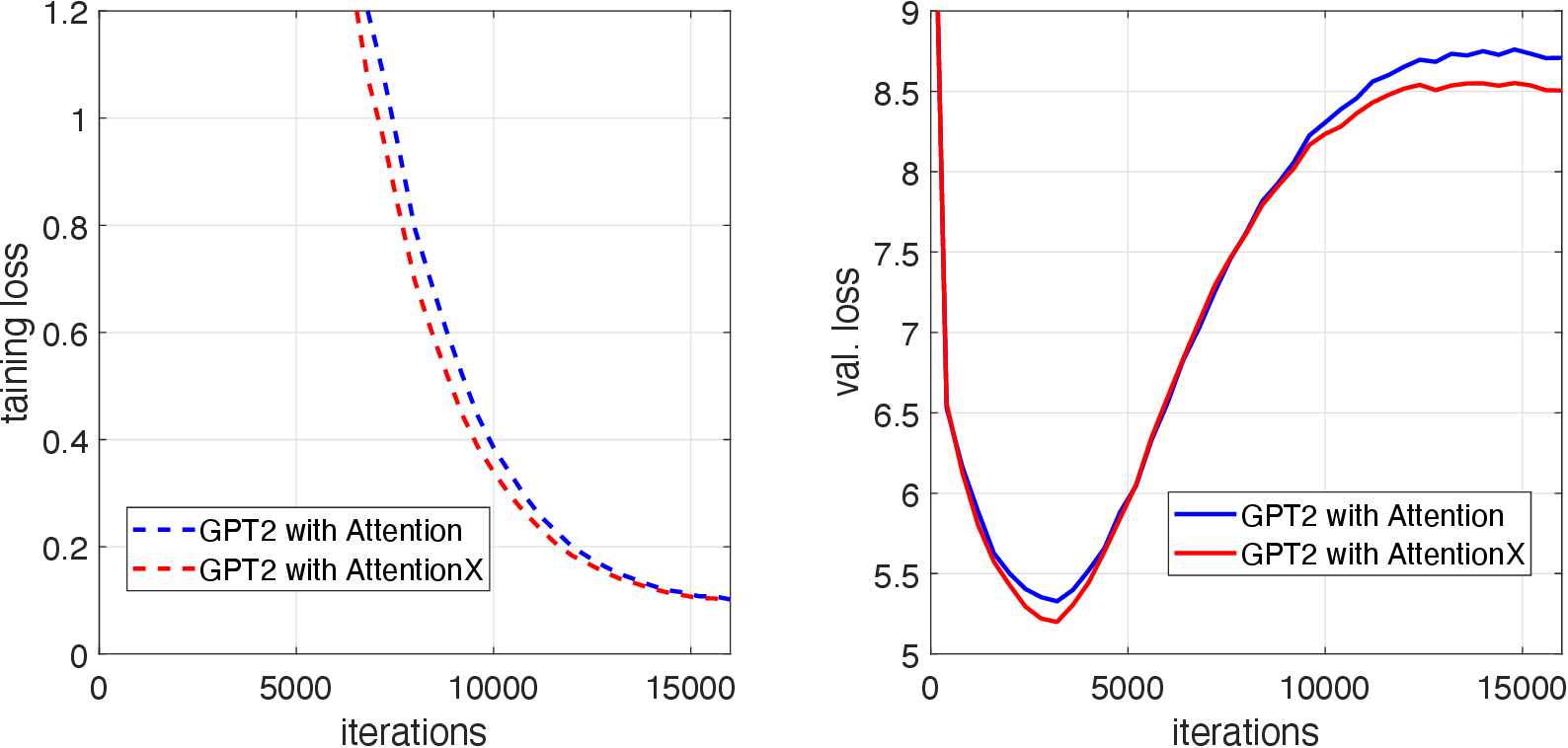}
\vspace*{-0.25cm}
\caption{\footnotesize{ Performance comparison when training GPT2. } }
\label{fig:GPT2}
\vspace*{-0.3cm}
\end{figure}

\subsection{On training nano-GPT2}

In this experiment, we consider training nano-GPT2 by using the dataset of openwebtext. To save training time, we only took a small subset from the entire training dataset when training the model. The parameter $\gamma$ in attentionX  was set to $\gamma=3$.

Fig.~\ref{fig:GPT2} summarizes the training and validation curves for using either Attention or AttentionX. It is clear from the plots that AttentionX makes the training procedure slightly faster. On the other hand, the resulting validation curve with  AttentionX is slightly better than with Attention. The above results are consistent with those of Table.~\ref{tab:AttentionX_ViT_small} for training ViT-small.

%\begin{figure}[h!]
%\centering
%\includegraphics[width=110mm]%{ViT_compare.eps}
%\vspace*{-0.25cm}
%\caption{\footnotesize{ Performance comparison when training ViT small. } }
%\label{fig:VIT_small}
%\vspace*{-0.3cm}
%\end{figure}

%\begin{figure}[t!]
%\centering
%\includegraphics[width=140mm]{image_editing_BDIA.pdf}
%\vspace*{-0.2cm}
%\caption{Image editing via BDIA-DDIM.}
%\label{fig:BDIA_DDIM}
%\vspace*{-0.0cm}
%\end{figure}

\vspace{-2mm}
\section{Conclusions}
\vspace{-2mm}

In this work, we have proposed AttentionX to replace Attention in transformer from a distributed optimization perspective. In particular, we first identify similarity between the update expressions of PDMM and the classical Attention-FFN framework in transformer. The Attention operation can be viewed as information gathering (corresponding to message passing in PDMM) from neighbouring tokens while the FFN operation can be taken as local information fusion (corresponding to the node-oriented optimization in PDMM). Inspired by PDMM that exploits the consensus discrepancy in its update expressions, we also utilize the consensus discrepancy in the form of the difference between the scaled weighted summation of $V$-representations and the  $V$-representations themselves when designing AttentionX. One additional learnable parameters are needed AttentionX. Experiments on ViT-small and nano-GPT2 show that AttentionX leads to better validation performance.
%{\small
%bibliographystyle{abbrv}
%bibliography{sigProcessing}
%}

%{\small
%\bibliographystyle{abbrv}
%\bibliography{article,sigProcessing}
%}

%%%%%%%%%%%%%%%%%%%%%%%%%%%%%%%%%%%%%%%%%%%%%%%%%%%%%%%%%%%%

\newpage

\appendix

\appendix
\onecolumn

\end{document}